\documentclass{article}
\usepackage{times}
\usepackage{epsfig}
\usepackage{graphicx}
\usepackage{amsmath}
\usepackage{amssymb}
\usepackage{bbm, dsfont}
\usepackage{color}
\usepackage{cite}
\usepackage[numbers,sort]{natbib}
\usepackage{amsmath}
\usepackage[margin=1.5in]{geometry}

\usepackage{graphicx}
\usepackage{comment}
\usepackage[utf8]{inputenc} % allow utf-8 input
\usepackage{url}            % simple URL typesetting
\usepackage{booktabs}       % professional-quality tables
\usepackage{amsfonts}       % blackboard math symbols
\usepackage{color}
\usepackage[ruled]{algorithm2e} 

\usepackage{algorithmic}
\usepackage{authblk}
 \usepackage{relsize}
\DeclareMathOperator*{\argmax}{arg\,max}
\DeclareMathOperator*{\argmin}{arg\,min}
\begin{document}
%\author{Yiming Xu, Diego Klabjan}
\author[1]{Yiming Xu}
\author[2]{Diego Klabjan}
\affil[1]{Department of Statistics, Northwestern University}
\affil[2]{Department of Industrial Engineering and Management Sciences, Northwestern University}
\title{Open Set Domain Adaptation by Extreme Value Theory}
\maketitle

\begin{abstract}
Common domain adaptation techniques assume that the source domain and the target domain share an identical label space, which is problematic since when target samples are unlabeled we have no knowledge on whether the two domains share the same label space. When this is not the case, the existing methods fail to perform well because the additional unknown classes are also matched with the source domain during adaptation. In this paper, we tackle the open set domain adaptation problem under the assumption that the source and the target label spaces only partially overlap, and the task becomes when the unknown classes exist, how to detect the target unknown classes and avoid aligning them with the source domain. We propose to utilize an instance-level reweighting strategy for domain adaptation where the weights indicate the likelihood of a sample belonging to known classes and to model the tail of the entropy distribution with Extreme Value Theory for unknown class detection. Experiments on conventional domain adaptation datasets show that the proposed method outperforms the state-of-the-art models.

\end{abstract}

\section{Introduction}
Recently, deep learning techniques have drawn a lot of attention in both academia and industry, due to their astounding performance in fields such as computer vision and natural language processing \cite{alexnet,vgg,resnet,resnext,gru,counting,pythia3,bottomup,fasterrcnn,Detectron}. However, one major disadvantage of deep learning is that neural networks generally require a large amount of training data to converge to the right solution and generalize. When the training data are insufficient, the model performance is usually adversely affected. Sometimes even if the training data are sufficient, the domain gap, i.e. the difference between data distributions, between the source domain (the data we train model on) and the target domain (the desired target task) may still contribute to low generalizability. This is because in conventional machine learning tasks, we usually assume that the training data distribution is the same as the testing data distribution. However, in real world, testing data are uncontrollable, and thus the difference between the source and the target domain can be substantial, which results in the overfitting problem, that is, the model does not generalize well to the testing set. 

In order to reduce the domain gap and better utilize the source domain knowledge, domain adaptation techniques have been proposed to resolve the issue. Domain adaptation assumes that the source domain has sufficient amount of labeled data to train a good model, while the desired target domain has insufficient amount of data to train the model. Note that in most domain adaptation tasks, the target domain does not have labeled data at all, which is also known as unsupervised domain adaptation. Domain adaptation methods leverage the knowledge from the source domain with sufficient labeled data to learn a model that works well in the target domain with insufficient or no labeled data. Typically, domain adaptation methods reduce the domain gap by diminishing the divergence between the label-rich source and the label-scarce target domain \cite{DA1,DA2,semida1,semida2,semida3,mmacm,adda,wdgrl,dan,dann,SE,MCD}. A significant drawback of most of the existing works is that they assume the source label space and the target label space are identical, i.e. the target classes are assumed to have appeared during the training process and the training classes do not contain classes that are not in the target domain. This is also known as closed set domain adaptation (CSDA), which is infrequent in real-world scenarios since it is hard to guarantee that the target domain classes are the same as the source domain classes. Aligning the source and the target domain by brute force when their label spaces are different is extremely detrimental to the model's generalizability, which is also known as the negative transfer phenomenon, because the CSDA methods would try to align the additional unknown classes as well during adaptation.

To handle the domain adaptation tasks without assuming identical label spaces, open set domain adaptation (OSDA) methods were proposed to first detect the irrelevant or unknown samples and then avoid adaptation on the unknown samples and perform domain adaptation only on the known classes \cite{ATI,OSBP,KASE,DFRODA}, which can be achieved by forcing the model to learn a clear boundary between known classes and unknown classes and a clear boundary within the known classes. After adaptation, the model is applied to the target samples as follows: either a target sample is classified as a class among the known classes, or rejected as an unknown class. However, all of the conventional OSDA methods employ a rejection threshold hyperparameter, where if a score or statistic for a sample is higher than the threshold, then the sample will be rejected as an unknown class and discarded during adaptation, and thus the model's sensitivity is largely affected by this rejection threshold.

In our work, we handle the existing issue in OSDA by 1) utilizing an entropy-based instance-level reweighting strategy and 2) extreme value theory (EVT) which has proven to be useful in many classification tasks due to its ability to model extreme values \cite{evt1,evt2,evt3,evt4,evt5}. We propose to use entropy of probability distribution of a sample to measure the likelihood it belongs to unknown classes. That is, the higher the entropy is, the more likely the sample belongs to the unknown classes, because the model is more uncertain regarding the prediction. We utilize the entropy values to construct an instance-level weight for domain adaptation, instead of setting a hard threshold. In this way, every sample is taken into account during the adaptation process, and a sample with high entropy should be focused less to avoid the negative transfer problem. In inference, we model the tail probability of the entropy distribution by fitting a generalized extreme value (GEV) distribution, and we use the cumulative distribution function (CDF) score to indicate if a sample belongs to unknown or known classes. Experimental results on three conventional domain adaptation datasets show outperformance over both state-of-the-art CSDA and OSDA benchmarks.

We claim the following key contributions. First, we model the tail of entropy distributions with EVT to detect and reject unknown classes. Second, the entropy-based instance-level weighting strategy avoids setting a hard threshold manually and thus is more robust and stable. Third, we have done extensive experiments on three conventional datasets in domain adaptation, which show that our model outperforms benchmarks by a significant margin.

The rest of the paper is organized as follows. In Section 2, we discuss the related literature and methods. In Section 3, we explain our OSDA method in details. In Section 4, we conduct experiments to validate the effectiveness of the proposed method and in Section 5, we conclude the paper by reiterating the main contributions.

\section{Related Work}

Existing works on CSDA typically try to diminish the domain gap by minimizing a statistical divergence between two domains for adaptation or by an adversarial approach. MCD \cite{MCD} utilize two task-specific classifiers to detect the target domain samples which are far from the source domain by maximizing the two classifiers' inconsistency regarding predictions. Based on the mean teacher model \cite{meanteacher} that is originally proposed for semi-supervised learning tasks, the self-ensemble network \cite{SE} is proposed to calculate the exponentially moving average of the student model weights and it assigns the weights to the teacher model to reduce the domain gap. There are also a plethora of works on adversarial learning to reduce domain gap. DANN \cite{dann} reduces the domain gap by introducing a domain discriminator during the training process to discriminate between domains where the domain discriminator is optimized by a specially designed gradient reversal layer. ADDA \cite{adda} incorporates adversarial training to reduce the domain gap by the discriminator to distinguish across domains. The soft labeling method \cite{DA1} utilizes a soft cross-entropy loss function to optimize for the domain invariance and thus aligning the source and target domains. WDGRL \cite{wdgrl} is proposed to minimize the Wasserstein distance across domains adversarially for learning domain invariance. CADA \cite{cada} aligns the joint distribution for labels and features across two domains by exploring classifier predictions for adversarial domain adaptation. Our proposed method differs from these prior works in that we focus on the practical OSDA setting while these prior works assume an identical label space across domains, and thus they cannot recognize the unknown classes samples included during domain adaptation. Our method is able to detect the unknown classes by instance-level weights based on entropy to avoid negative transfer.

There are also several existing works on OSDA to address the issue of unknown classes. STA \cite{SeparateToAdapt} is proposed to develop a coarse-to-fine progressive separation method for unknown and known classes. OSBP \cite{OSBP} forces the generator to either match target samples with source known classes or ignore them in adaptation as unknown samples by adversarial training. Based on the work of SE \cite{SE} in CSDA, \cite{KASE} modifies the adaptation loss and then aligns the target domain with the source domain to address the unknown classes using weights calculated by entropy values. By factorization and joint separation, D-FRODA \cite{DFRODA} represents the source and target classes with a shared embedding space for domain adaptation. ATI \cite{ATI} is also proposed to detect the target samples that potentially belong to the known classes by learning a mapping from the source domain to the target domain. We see the difference between our proposed work and the prior works in the utilization of a hard threshold to recognize and reject the unknown samples, and thus the model accuracy largely depends on how the threshold is set. The proposed method utilizes entropy to indicate the likelihood of unknown classes and it constructs a soft instance-level weight. We further fit an EVT model on the tail of the entropy distribution to detect unknown classes. Therefore, by avoiding manual thresholding and introducing EVT to detect unknown classes, the proposed method is more robust and suitable in the OSDA setting, which is also validated in the experimental results.

\section{Methodology}

In this section, $K$ denotes the number of known classes; $x^s,x^t$ denote a source, target sample, respectively; $y^s$ denotes a source label; $X^s,X^t$ denote the source, target sample distribution, respectively; $D^s,D^t$ denote the source, target domain, respectively; $g$ denotes a feature extractor; $clf$ denotes a $K$-way classifier to classify samples into one of the $K$ known classes; and $clf^d$ denotes the binary domain classifier to classify samples into source or target domains.

Even though all samples in the source domain are labeled, in one loss component we assume a split of known classes to $S$ (the classes of interest) and $C$ (the irrelevant classes), where $S \cap C =\phi, S \cup C = \{1,2,...,K\} $, and we treat the classes in $C$ as unknown. To this end, we denote by $X^s_U$ as the sample distribution of the samples corresponding to the classes in $C$ (unknown-class distribution). This is a known concept, e.g. \cite{KASE,OSBP}.

\subsection{Loss functions}
\indent

\textbf{Entropy Weighted Domain Adversarial Training:} In our work, we propose a novel loss to train the model in an adversarial manner by a domain classifier $clf^d$ to distinguish samples from the source and target domains, where $clf^d$ outputs a probability indicating the likelihood of belonging to the target domain, with instance-level weights on target samples to avoid negative transfer in OSDA. The domain classifier $clf^d$ is optimized based on a binary cross-entropy loss. We train the feature extractor $g$ to maximize the domain classification loss and the domain classifier to minimize the domain classification loss.

The domain classification loss is calculated by

$$L_d =  E_{x^s \sim X^s} log(clf^d(g(x^s))) + E_{x^t \sim X^t} [w(x^t)\cdot log(1-clf^d(g(x^t)))]$$
where
$$
w(x^t)=\frac{1}{Z} exp{\left(-\sum^{K}_{c=1} clf(g(x^t))_c \log clf(g(x^t))_c\right)}.
$$
Here $Z$ is the partitioning function such that $\sum w\left(x^t\right)=1$.

The feature extractor $g$ is trained to maximize $L^d$ so that the embeddings from both domains are similar and the domain classifier $clf^d$ is trained to minimize $L^d$ for a good classification performance. Classifier $clf$ is trained together with a different loss component which is discussed later. We make the feature extractor fool the domain classifier and the domain classifier learns how to perform better from the feature extractor in the adversarial manner.

%$$L_1 = \frac{\sum_{x^t \sim X^t} w*|| g(x^t) - sourceEmb ||_2}{n^t * \sum w(x^t)}$$
%$$L_1 = \sum log (wa /\sum w)=\sum log(wa) - nlog(\sum w)$$
%$$dL/d\theta = \sum \frac{1}{wa}* \frac{d(wa)}{d\theta} - n\frac{1}{\sum w} d \sum w/d\theta=\sum \frac{1}{wa} \frac{d(wa)}{d\theta} - n\frac{\sum d w/d\theta}{\sum w} $$

\textbf{Entropy Maximization for Source Unknown Classes:} The entropy of classifier predictions is high if the sample is from unknown classes and vice versa, because the classifier $clf$ is optimized to minimize the cross-entropy loss on source known classes. Therefore, the known classes predictions are close to one-hot vectors with low entropy. Based on this observation, the unknown classes detection task can be tackled by separating the known classes and unknown classes by their entropy values. In order to force the classifier to predict unknown classes with high entropy, we penalize the model if the prediction is different from the uniform distribution for a source unknown sample, because we want to force the unknown classes predictions as uncertain as possible. The loss function component is calculated as

$$
    L_e = -E_{x^s \sim X^s_{U}} [-\sum^{K}_{c=1} clf(g(x^s))_c \log clf(g(x^s))_c].
$$
By maximizing the source unknown entropy and minimizing the source known entropy, we can further separate the known and unknown classes with a clear boundary based on their entropy values.

\textbf{Total Loss:} Let us denote $\theta^g$ as the parameters of the feature extractor $g$, $\theta^c$ as the parameters of the classifier $clf$ and $\theta^d$ as the parameters of the domain classifier $clf^d$. The total loss to minimize is calculated as

$$L = - \lambda_d L_d + \lambda_e L_e + \lambda_c L_c$$                    
where $L_c=E_{(x^s,y^s) \sim D^s} CE(clf(g(x^s)),y^s)$ is the conventional cross-entropy classification loss on source known classes and $\lambda_d, \lambda_e, \lambda_c$ are weights for the loss components.

To force the domain classifier to minimize the adversarial loss and the feature extractor to maximize the adversarial loss, we seek a saddle point $\hat{\theta}^g, \hat{\theta}^c, \hat{\theta}^d$ of $L$ satisfying the conditions:
\begin{eqnarray}
\hat{\theta}^g, \hat{\theta}^c& \in &\argmin_{\theta^g,\theta^c} L(\theta^g, \theta^c, \hat{\theta}^d)\nonumber\\
\hat{\theta}^d& \in &\argmax_{\theta^d} L(\hat{\theta}^g, \hat{\theta}^c, \theta^d).
\end{eqnarray}

On a saddle point, $\theta^d$ minimize the domain classification loss $L_d$,  $\theta^c$ minimize the conventional classification loss $L_c$, $\theta^g$ maximize the domain classification loss (thus the feature divergence is minimized across the two domains) and $\theta^g$ minimize the classification loss $L_c$ (thus the features are discriminative). 

In the total loss $L$, in order to align domains more effectively in OSDA, we propose a novel loss component $L_d$ with domain discriminator and instance-level weights to reduce the domain gap adversarially. Components $L_e$ and $L_c$ already appear in the existing works such as \cite{KASE, OSBP}.

\textbf{Target Sample Classification:} We propose to fit GEV using the tail of the entropy distribution of source samples. The probability density function for GEV is calculated as
$$PDF(x; l, s, c)= t(x; l, s, c) ^{1 + c} \cdot exp(
  - t(x; l, s, c) ) / s$$
  where   $$ t(x; l, s, c) = (1 + c \cdot (x - l) / s) )^{-1 / c}$$
and the CDF is calculated as $$GEV(x; l, s, c) = exp(-t(x; l, s, c)),$$
where $l, s, c$ are parameters associated with the distribution.

After fitting GEV using the source samples' entropy values based on the trained $clf$, for target samples, we calculate the CDF based on the source sample GEV. If the CDF value is higher than 0.5, then the sample is classified as an unknown sample; otherwise it is input into the known-class classifier $clf$ to be predicted into one of the $K$ known classes. We abbreviate our overall methodology as ADAGEV (Adversarial Domain Adaptation with GEV).

\subsection{Training  strategy}

We train our model to minimize loss function $L$. Since we want the feature extractor to maximize the adversarial loss, we utilize the gradient reversal layer \cite{dann} as the first layer of the domain discriminator $clf^d$ such that during forward propagation, the layer is essentially an identity function, while in backward propagation, the gradient is multiplied by $-1$. In this way, the total loss function can be jointly optimized in one single step instead of alternate training, i.e. optimizing each loss component individually step by step. There are two advantages of avoiding alternate training: 1. lower computational costs, as fewer gradient iterations are calculated and 2. fewer hyperparameters to tune, e.g. optimizing one component for $N$ times and optimizing another component for $M$ times are absent, etc.

To estimate the partitioning function, we have tried the following approaches: 1. using the same batch of data (as the batch sampled to calculate the numerator of $w(x)$) to calculate $\sum w(x)$; 2. using a different batch of data of the same size $B$ to calculate $\sum w(x)$; 3. using the sum of the values of the first two approaches (sum of weights of the $2B$ samples). We find that in practice, the first approach achieves overall the best performance with the lowest computational cost, and thus it is used in ADAGEV. The entire algorithm is exhibited in Algorithm 1.

\begin{algorithm}[h!]
 \For{t = 1,...,T}{
 Draw $B$ samples and corresponding labels $\tilde{x}^s,\tilde{y}^s$ from the source known-class data\;
 Draw $B$ samples $\tilde{x}^s_U$ from the source unknown-class data\;
 Draw $B$ samples $\tilde{x}^t$ from the target data\;

  %Randomly draw $B$ samples from the source domain, and calculate $w(x)$\;
  Estimate the partitioning function $Z$ by $z=\mathlarger{\sum}_{x \in \tilde{x}^t} w(x)$\ \;
 Update parameters by a gradient iteration: $\Theta^t$$=\Theta^t-\alpha\nabla L(\tilde{x}^s,\tilde{y}^s,\tilde{x}^s_U,\tilde{x}^t,\Theta^t)$\;
 }
 
 Fit GEV using the source sample's entropy values\;
  \For{$x^t$ in the target data}{

 \uIf{CDF of entropy of $clf(g(x^t)) > 0.5$}{
    $x^t$ belongs to unknown-class\;
  }
\Else{
    Calculate $\argmax clf(g(x^t))$ which is the predicted known-class label for $x^t$\;
  }
%\ENDIF
 }
 
 \caption{ADAGEV }
\end{algorithm}

\section{Experimental results}

We evaluate ADAGEV on three conventional benchmark datasets in domain adaptation: Digits, Office-31 \cite{office31} and VisDA \cite{visda2017}. In the experiments, the source domain samples are labeled and the target domain samples are not labeled, and the task is to either classify target samples into one of the known classes or reject as unknown. We use the conventional OSDA metrics OS and OS* as performance measures, which denote the mean accuracy for all $K+1$ classes (all known classes and the additional unknown class) and the mean accuracy for the $K$ known classes, respectively. The hyperparameters are set as $\lambda_d=0.5, \lambda_e=1, \lambda_c=1$ (discussed later). The benchmark state-of-the-art methods are discussed in Section 2, and the accuracies are obtained from the corresponding papers except \cite{KASE}, which only reported the OS accuracy. To complete the comparison for OS*, we ran the source code of \cite{KASE} and after finetuning we are able to achieve a slightly better performance for OS than the original paper, and both the OS and OS* accuracies are reported in our experiments. We discuss the network architectures for each experiment in the following subsections. For a fair comparison, the same network architecture is used in ADAGEV as in the benchmark algorithms (which all use the same architecture). The stopping criteria are also discussed in the following subsections.
% except for \cite{KASE}, which only reported the OS accuracy. To complete the comparison for OS*, we ran the source code and we are able to achieve a slightly better performance than the original paper, which is reported in our experiments.
\subsection{Digits Experiments}

In the Digits experiments, we follow the conventions in OSDA to use USPS \cite{USPS}, SVHN \cite{SVHN} and MNIST to conduct evaluations, and we perform domain adaptation from SVHN to MNIST, USPS to MNIST and MNIST to USPS. Note that the USPS to MNIST and the MNIST to USPS tasks are easier than the SVHN to MNIST task, since MNIST and USPS both contain 2-dimensional black-and-white images and are hand-written digits, while SVHN contains 3-dimensional RGB images for real-world house numbers which are part of Google Street View images. Therefore, the domain gap between SVHN to MNIST is significantly larger than in the other two experiments. For consistency, we use digits 0, 1, 2, 3 as known classes, 4, 5, 6 as source unknown classes and 7, 8, 9 as target unknown classes, following the conventional protocols. Also following the existing works, we use the same CNN architectures (details can be found in the appendix of \cite{OSBP}) for a fair comparison. We compare ADAGEV with the state-of-the-art methods in both CSDA and OSDA which are discussed in Section 2. We report accuracies for ADAGEV after training for 20 epochs. For benchmarks, they either not mention the stopping criteria or use 200 epoches. 

\begin{table}[h!]%\fontsize{7}{9}\selectfont
%\vspace{-0.1in}
\centering
\caption{Accuracy (in \%) on Digits dataset (best in bold). `AVG' denotes the average across all datasets.} %Breakdown scores are obtained from the online submission system. 
  \begin{tabular}{l |c c| c c |c c|c c}
  \hline
    Method & \multicolumn{2}{c|}{\textbf{S-M}}&\multicolumn{2}{c|}{\textbf{U-M}}&\multicolumn{2}{c|}{\textbf{M-U}}&\multicolumn{2}{c}{\textbf{AVG}}\\ 
    & OS&OS*&OS&OS*&OS&OS*&OS&OS* \\ \hline
Source only &58.5& 63.5& 82.3& 83.9&82.0&84.1 &74.2 & 77.2\\
DAN \cite{dan} &66.2& 67.0& 86.9&88.0 &89.1& 90.5&80.7 &81.8\\
DANN \cite{dann} &66.8&67.4 &89.2& 88.9&88.9& 89.9&81.6 &82.0 \\ 
OSBP \cite{OSBP} &62.2& 63.9 &94.8&94.3 &92.7& 93.2&83.2 & 83.8\\ 
STA \cite{SeparateToAdapt} &65.2&65.9 &94.7& 94.5&93.3& 94.1&84.4 &84.8 \\ 
KASE \cite{KASE} &67.2 &68.1 & 94.7 & 95.1 &93.6 &94.5 &85.2 &85.9\\ \hline
%Ours w/o entropy weights&66.7&92.7&86.8&82.1 \\ \hline
ADAGEV & \textbf{87.9} &\textbf{89.2} &\textbf{95.4}& \textbf{96.3}&\textbf{95.2}&\textbf{96.5} &\textbf{92.8} & \textbf{94.0}\\ 
Relative Imp & \textbf{30.8\%} &\textbf{31.0\%} &\textbf{0.7\%}& \textbf{1.3\%}&\textbf{1.7\%}&\textbf{2.1\%} &\textbf{8.9\%} & \textbf{9.4\%}\\ \hline
  \end{tabular}
  \label{tab:digits}
  %\vspace{-0.2in}
\end{table}

Table \ref{tab:digits} shows the overall performance comparison, where the model trained on source only performs the worst, which is expected because it does not align the two domains at all. The CSDA benchmarks DAN and DANN perform slightly better than training only on source data, because they reduce domain misalignment, but they do not reject unknown classes in domain adaptation. The OSDA methods OSBP, STA and KASE outperform the CSDA methods but only improve by a small margin (1.6\%, 2.8\%, 3.6\%, respectively), from which we observe the difficulty in the OSDA scenario. ADAGEV achieves 92.8\% OS accuracy, outperforming the previous state-of-the-art OSDA method by 9.6\%, 8.4\% and 7.6\% respectively. The method also outperforms the CSDA benchmarks by 12.1\% and 11.2\%. In the most difficult task S-M, our method is able to achieve more than 20\% improvement on both OS and OS* accuracies, which again validates the effectiveness of the proposed method in the OSDA scenario. We also experiment on 5 different random seeds and the standard deviations for average OS and OS* are 0.49 and 0.37, indicating the performance is robust and stable. We also observe that comparing the OS and OS* metrics, OS* is always higher than OS for the same method, meaning the unknown class accuracy is lower than the overall accuracy on known classes only. This points to the future research direction in the OSDA area that the unknown classes should be addressed more on their detection and separation. Our relative OS improvements regarding benchmarks are 25.1\%, 15.0\%, 13.7\%, 11.5\%, 10.0\%, 8.92\% which are substantial improvements.

\begin{figure}[h]
    \includegraphics[width=0.7\textwidth]{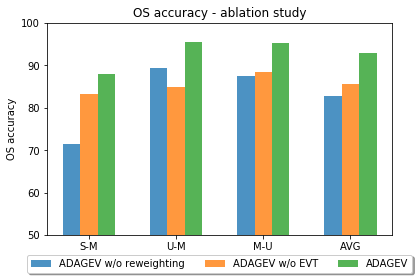}
    \centering
  \caption{OS (in \%) on Digits dataset without specific parts of our model. `AVG’ denotes the average across all datasets. Note that the y-axis starts from 50 instead of 0.}
  \centering
  \label{ablation}
  \end{figure}

\textbf{Ablation study}
In order to have a better understanding on what component in our model contributes most to the performance, we conduct ablation studies by removing components from our model. The ablation study results are exhibited in Figure \ref{ablation}, where only the OS accuracy is shown for simplicity since the OS* accuracy shows a very similar performance, i.e. the OS* accuracy is consistently higher than the OS accuracy by 1\% $\sim$ 2\%. Note that in the second experiment (i.e. w/o EVT) we replace EVT by a binary classifier to classify target samples into known/unknown classes where the classifier is trained using source known and unknown samples.

From Figure \ref{ablation} we observe that removing the soft reweighting strategy contributes most to the accuracy drop, i.e. the average OS drops by 10.0\% without entropy-based reweighting. Removing the EVT component is also detrimental to the performance, where the average OS drops by 7.3\% without EVT. Including both reweighting and EVT in ADAGEV achieves the new state-of-the-art performance in OSDA.

\begin{figure}[h]
    \includegraphics[width=0.7\textwidth]{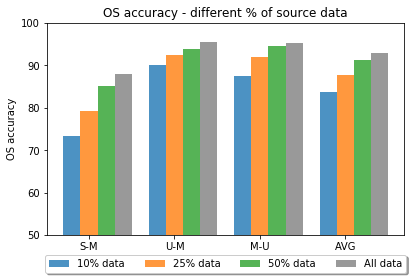}
    \centering
  \caption{Performance of ADAGEV on Digits dataset with different percentage of source data. Note that the y-axis starts from 50 instead of 0.}
  \centering
  \label{digits_fewer}
  \end{figure}

\textbf{Experiments with less source data}
In domain adaptation tasks, it is usually assumed that the source domain contains sufficient amount of labeled data. In order to see how our method performs with less source data, we experiment on limiting the source data to 10\%, 25\% and 50\%. From Figure \ref{digits_fewer}, both the OS and OS* accuracies increase when we have more source data. The model trained with 10\% data achieves comparable performance with the benchmarks, and the model trained with 25\% source data already outperforms the previous state-of-the-art OSDA method. This validates that ADAGEV shows robust performance even when less source data are presented. 

\begin{figure}[h]
    \includegraphics[width=0.6\textwidth]{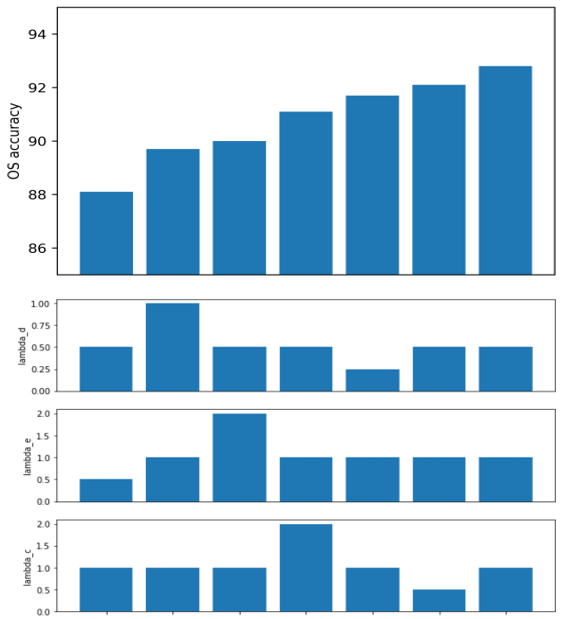}
    \centering
  \caption{Model performance on varying loss function weights on Digits dataset.  }
  \centering
  \label{digits_loss_weights}
  \end{figure}

\textbf{Sensitivity on hyperparameters}
We also experiment on varying the loss function weight hyperparameters to observe their sensitivity. The results are shown in Figure \ref{digits_loss_weights}, where in general, the hyperparameter for source classification loss $\lambda_c$ is less sensitive in that varying its value results in the least accuracy changes, which we postulate is because the source knowledge can already be learned well for a small weight. The hyperparameters for domain discriminator loss $\lambda_d$ and unknown entropy maximization loss $\lambda_e$ are more sensitive to the changes where either increasing or decreasing the values results in an accuracy drop. The combination $\lambda_d=0.5, \lambda_e=1.0, \lambda_c=1.0$ which provides the best accuracy is used as the final hyperparameters across all datasets.

\subsection{Office-31 Experiments}

We also experiment on the Office-31 dataset which is another frequently used domain adaptation dataset containing three domains: Amazon, DSLR and Webcam. The following 6 domain adaptation tasks are created: Amazon to DSLR, DSLR to Amazon, Amazon to Webcam, Webcam to Amazon, DSLR to Webcam and Webcam to DSLR. Each domain contains 31 classes, and the images are either collected directly from www.amazon.com or they are office environment images taken with different lighting, etc. using a webcam or a digital single-lens reflex (DSLR) camera. Therefore, the domain gap between DSLR and Webcam is slightly smaller. Following the same experimental setup as in previous works, we set the 10 common classes from the Office-Caltech dataset \cite{semida1} as the known classes, and the first 10 remaining classes in the alphabetical order are set as the unknown classes for the source domain and the last 11 classes are set as the unknown classes for the target domain. We use the same CNN model AlexNet \cite{alexnet} as in previous works across all methods for fair comparison. Following the conventions in \cite{OSBP}, we report accuracies after training for 50 epoches. The experimental results are presented in Table \ref{tab:office}.

\begin{center}
\begin{table*}[h!]%\fontsize{7}{9}\selectfont
%\centering
\footnotesize
\caption{Accuracy (in \%) on Office-31 dataset (best in bold). `AVG' denotes the average across all datasets. `/' denotes the number is unavailable because the cited paper does not include such experiment.  } %Breakdown scores are obtained from the online submission system. 
  \begin{tabular}{l | c c| c c| c c| c c|  c c| c c |c c}
  \hline
     Method & \multicolumn{2}{c|}{\textbf{A-D}}&\multicolumn{2}{c|}{\textbf{A-W}}&\multicolumn{2}{c|}{\textbf{D-A}}&\multicolumn{2}{c|}{\textbf{D-W}}&\multicolumn{2}{c|}{\textbf{W-A}}&\multicolumn{2}{c|}{\textbf{W-D}}&\multicolumn{2}{c}{\textbf{AVG}}\\ 
         & OS&OS*&OS&OS*&OS&OS*&OS&OS* & OS&OS*&OS&OS*&OS&OS* \\ \hline

Source only &68.5& 71.8 & 59.6& 63.2 & 54.8& 58.1 & 88.2& 90.5 & 49.8& 53.2 & 92.6& 92.9 & 68.9&71.6\\
DAN \cite{dan} &78.7  & 79.5&73.4& 74.2 & 59.6& 62.1 & 87.8  &88.6 &62.2& 64.5 & 96.5  &96.8 &76.3&77.6\\
DANN \cite{dann} &79.7  & 80.5& 77.2  & 79.3& 55.2 &56.1 &  90.7  & 91.3& 65.7  &67.0 & 97.9  & 98.3& 78.0&78.8 \\ 
ATI \cite{ATI} &79.8 &79.2  &77.6& 76.5 & 71.3& 70.0&  93.5& 93.2 & 76.7& 76.5 & 98.3& 99.2 & 82.9 &82.4 \\ 
OSBP \cite{OSBP} &74.7& 76.4&  72.6& 73.1&  61.7&63.3 &  93.3&94.8 &  79.8& 82.5&  95.7&96.5 &  79.6&81.1 \\ 
D-FRODA \cite{DFRODA} &87.4& / &  78.1&/ &  73.6&/ &  94.4&/ &  77.1&/ &  98.5&/ &  84.9& / \\ 
KASE \cite{KASE} &86.6& 88.1&  80.1& 80.5&  78.6& 79.8&  95.4& 95.9&  82.2& 83.4&  98.9&  98.8& 87.0&87.8 \\ \hline
ADAGEV&\textbf{86.7}& \textbf{89.9} & \textbf{81.3}&\textbf{85.4} & \textbf{81.3}&\textbf{83.7} & \textbf{96.4}& \textbf{96.8}& \textbf{82.9}& \textbf{84.8}& \textbf{99.1}&\textbf{99.5} & \textbf{88.4} & \textbf{89.8}\\ 
Relative Imp&\textbf{0.1\%}& \textbf{2.0\%} & \textbf{1.5\%}&\textbf{6.1\%} & \textbf{3.4\%}&\textbf{4.9\%} & \textbf{1.0\%}& \textbf{0.9\%}& \textbf{0.9\%}& \textbf{1.7\%}& \textbf{0.2\%}&\textbf{0.7\%} & \textbf{1.6\%} & \textbf{2.3\%}\\ \hline

  \end{tabular}
  \label{tab:office}
\end{table*}
\end{center}
From Table \ref{tab:office}, the CSDA methods generally have accuracy below 80\% since they are unable to address the additional unknown classes during adaptation. The OSDA methods outperform the CSDA methods in this experiment with a large margin, where the previous state-of-the-art method KASE achieves 87.0\% OS accuracy and 87.8\% OS* accuracy, significantly outperforming other methods. ADAGEV shows 88.4\% OS and 89.8\% OS* accuracies, consistently beating the previous state-of-the-art method in all domain adaptation experiments for both known and unknown classes. Our relative OS improvements regarding benchmarks vary from 1.6\% to 28.3\% with the average being 11.5\%.
%are 28.3\%, 15.9\%, 13.3\%, 6.63\%, 11.1\%, 4.1\% and 1.6\%.

\subsection{VisDA Experiments}

The VisDA dataset \cite{visda2017} is a conventional but more difficult task in domain adaptation. The source domain contains synthetic images and the target domain contains real-world images. For fair comparison, we follow the conventions to set ``bicycle,'' ``bus,'' ``car,'' ``motorcycle,'' ``train,'' ``truck'' as 6 known categories. In alphabetical order, the first 3 remaining categories ``aeroplane,'' ``horse,'' ``knife'' are set as source unknown categories and the last 3 remaining categories ``person,'' ``plant,'' ``skateboard'' are set as target unknown categories. We report accuracies after training for 50 epoches. %Some of the source and target samples are shown in Figure~\ref{fig:osda_source_target_examples}.

\begin{table*}[h!]%\fontsize{7}{9}\selectfont
\centering
\caption{Accuracy (in \%) on VisDA-2017 (best in bold). `UNK' denotes the additional unknown class.} %Breakdown scores are obtained from the online submission system. 
  \begin{tabular}{l | c  c c c  c c c|c c}
  \hline
     Method &       \textbf{bicycle}&\textbf{bus}&\textbf{car}&\textbf{motorcycle}&\textbf{train}&\textbf{truck}&\textbf{UNK}&\textbf{OS}&\textbf{OS*}\\ \hline
Source only &42.9& 65.9& 58.9& 80.5& 80.2& 8.7& 10.1&49.6&56.2\\
DANN \cite{dann} & 46.1&  69.0& 56.2& 84.4& 82.8& 18.2& 52.0& 58.4&59.5 \\ 
OSBP \cite{OSBP} &51.3& 70.7& 37.3& 87.8& 77.3& 23.8& 88.1 & 62.3 &58.0 \\ \hline
ADAGEV & \textbf{56.0}& \textbf{71.2}&\textbf{59.1}&\textbf{88.4}&\textbf{83.0}&\textbf{25.0}&\textbf{88.3}&\textbf{67.3}&\textbf{63.8} \\ \hline

  \end{tabular}
  \label{tab:visda}
\end{table*}

From Table \ref{tab:visda}, the previous state-of-the-art methods DANN and OSBP consistently outperform the model trained only on source data, while ADAGEV is able to achieve the OS accuracy gains of 17.7\%, 8.9\%, 5.0\% and OS* accuracy gains of 7.6\%, 4.3\%, 5.8\% comparing with the benchmarks, from which the effectiveness of the proposed method is validated again even in this difficult task. Our relative OS improvements regarding benchmarks are 35.7\%, 15.2\% and 8.0\%.

\section{Conclusion}

We propose an open set domain adaptation method which models the tail of the entropy distributions using EVT and utilizes an instance-level reweighting strategy to detect and reject unknown samples. Experiments show that the method achieves the new state-of-the-art performance by beating the existing benchmarks by a large margin for three conventional domain adaptation datasets.

\bibliography{ref}

\begin{thebibliography}{10}
\providecommand{\url}[1]{\texttt{#1}}
\providecommand{\urlprefix}{URL }
\providecommand{\doi}[1]{https://doi.org/#1}

\bibitem{bottomup}
Anderson, P., He, X., Buehler, C., Teney, D., Johnson, M., Gould, S., Zhang,
  L.: Bottom-up and top-down attention for image captioning and visual question
  answering. In: CVPR (2017)

\bibitem{DFRODA}
Baktash, M., Faraki, M., Drummond, T., Salzmann, M.: Learning factorized
  representations for open-set domain adaptation. In: ICLR (2019)

\bibitem{evt5}
Bendale, A., Boult, T.: Towards open set deep networks. In: CVPR (2016)

\bibitem{ATI}
Busto, P.P., Gall, J.: Open set domain adaptation. In: ICCV (2017)

\bibitem{gru}
Cho, K., van Merri{\"e}nboer, B., Gulcehre, C., Bougares, F., Schwenk, H.,
  Bengio, Y.: Learning phrase representations using {RNN} encoder-decoder for
  statistical machine translation. In: EMNLP (2014)

\bibitem{evt3}
Dang, S., Cao, Z., Cui, Z., Pi, Y., Liu, N.: Open set incremental learning for
  automatic target recognition. In: IEEE T-GRS (2019)

\bibitem{SE}
French, G., Mackiewicz, M., Fisher, M.H.: Self-ensembling for visual domain
  adaptation. In: ICLR (2017)

\bibitem{dann}
Ganin, Y., Lempitsky, V.S.: Unsupervised domain adaptation by backpropagation.
  In: ICML (2015)

\bibitem{evt1}
Geng, C., Huang, S.J., Chen, S.: Recent advances in open set recognition: A
  survey. In: IEEE T-PAMI (2020)

\bibitem{Detectron}
Girshick, R., Radosavovic, I., Gkioxari, G., Doll\'{a}r, P., He, K.: Detectron.
  \url{https://github.com/facebookresearch/detectron} (2018)

\bibitem{semida1}
Gong, B., Shi, Y., Sha, F., Grauman, K.: Geodesic flow kernel for unsupervised
  domain adaptation. In: CVPR (2012)

\bibitem{semida2}
Guo, Y., Xiao, M.: Cross language text classification via subspace
  co-regularized multi-view learning. In: ICML (2012)

\bibitem{resnet}
He, K., Zhang, X., Ren, S., Sun, J.: Deep residual learning for image
  recognition. In: CVPR (2015)

\bibitem{DA1}
Hoffman, J., Tzeng, E., Darrell, T., Saenko, K.: Simultaneous deep transfer
  across domains and tasks. In: ICCV (2015)

\bibitem{DA2}
Koniusz, P., Tas, Y., Porikli, F.: Domain adaptation by mixture of alignments
  of second- or higher-order scatter tensors. In: CVPR (2017)

\bibitem{alexnet}
Krizhevsky, A., Sutskever, I., Hinton, G.E.: Imagenet classification with deep
  convolutional neural networks. In: NIPS (2012)

\bibitem{USPS}
Lecun, Y., Bottou, L., Bengio, Y., Haffner, P.: Gradient-based learning applied
  to document recognition. In: Proceedings of the IEEE (1998)

\bibitem{KASE}
Lian, Q., Li, W., Chen, L., Duan, L.: Known-class aware self-ensemble for open
  set domain adaptation. In: arXiv:1905.01068 (2019)

\bibitem{SeparateToAdapt}
Liu, H., Cao, Z., Long, M., Wang, J., Yang, Q.: Separate to adapt: Open set
  domain adaptation via progressive separation. In: CVPR (2019)

\bibitem{cada}
Long, M., Cao, Z., Wang, J., Jordan, M.I.: Conditional adversarial domain
  adaptation. In: NIPS (2018)

\bibitem{dan}
Long, M., Zhu, H., Wang, J., Jordan, M.I.: Unsupervised domain adaptation with
  residual transfer networks. In: NIPS (2016)

\bibitem{SVHN}
Netzer, Y., Wang, T., Coates, A., Bissacco, A., Wu, B., Ng, A.: Reading digits
  in natural images with unsupervised feature learning. In: NIPS (2011)

\bibitem{evt2}
Oza, P., Patel, V.: C2ae: Class conditioned auto-encoder for open-set
  recognition. In: CVPR (2019)

\bibitem{visda2017}
Peng, X., Usman, B., Kaushik, N., Hoffman, J., Wang, D., Saenko, K.: Visda: The
  visual domain adaptation challenge. In: arXiv:1710.06924 (2017)

\bibitem{mmacm}
Qi, F., Yang, X., Xu, C.: A unified framework for multimodal domain adaptation.
  In: ACM Multimedia (2018)

\bibitem{fasterrcnn}
Ren, S., He, K., Girshick, R., Sun, J.: Faster r-cnn: Towards real-time object
  detection with region proposal networks. IEEE T-PAMI  (2015)

\bibitem{office31}
Saenko, K., Kulis, B., Fritz, M., Darrell, T.: Adapting visual category models
  to new domains. In: ECCV (2010)

\bibitem{MCD}
Saito, K., Watanabe, K., Ushiku, Y., Harada, T.: Maximum classifier discrepancy
  for unsupervised domain adaptation. In: CVPR (2018)

\bibitem{OSBP}
Saito, K., Yamamoto, S., Ushiku, Y., Harada, T.: Open set domain adaptation by
  backpropagation. In: ECCV (2018)

\bibitem{wdgrl}
Shen, J., Qu, Y., Zhang, W., Yu, Y.: Wasserstein distance guided representation
  learning for domain adaptation. In: AAAI (2017)

\bibitem{vgg}
Simonyan, K., Zisserman, A.: Very deep convolutional networks for large-scale
  image recognition. In: arXiv:1409.1556 (2014)

\bibitem{pythia3}
Singh, A., Natarajan, V., Shah, M., Jiang, Y., Chen, X., Batra, D., Parikh, D.,
  Rohrbach, M.: Towards {VQA} models that can read. In: CVPR (2019)

\bibitem{meanteacher}
Tarvainen, A., Valpola, H.: Mean teachers are better role models:
  Weight-averaged consistency targets improve semi-supervised deep learning
  results. In: ICLR (2017)

\bibitem{adda}
Tzeng, E., Hoffman, J., Saenko, K., Darrell, T.: Adversarial discriminative
  domain adaptation. In: CVPR (2017)

\bibitem{resnext}
Xie, S., Girshick, R.B., Doll{\'{a}}r, P., Tu, Z., He, K.: Aggregated residual
  transformations for deep neural networks. In: CVPR (2017)

\bibitem{semida3}
Yao, T., Pan, Y., Ngo, C.W., Li, H., Mei, T.: Semi-supervised domain adaptation
  with subspace learning for visual recognition. In: CVPR (2015)

\bibitem{evt4}
Zhang, H., Patel, V.M.: Sparse representation-based open set recognition. In:
  IEEE T-PAMI (2017)

\bibitem{counting}
Zhang, Y., Hare, J., Prügel-Bennett, A.: Learning to count objects in natural
  images for visual question answering. In: ICML (2018)

\end{thebibliography}
\bibliographystyle{splncs04}

\end{document}